# 4DPV: 4D Pet from Videos by Coarse-to-Fine Non-Rigid Radiance Fields

Sergio M. de Paco and Antonio Agudo

Institut de Robòtica i Informàtica Industrial, CSIC-UPC, Barcelona, Spain
{smontoya,aagudo}@iri.upc.edu

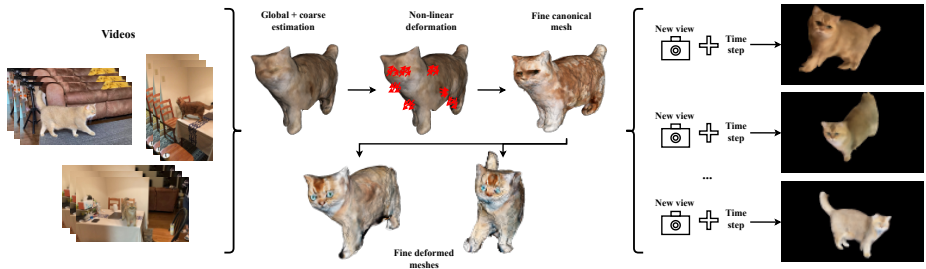

**Fig. 1: 4D Shapes from multiple videos.** Given multiple monocular videos of an unknown and dynamic object shape with non-controlled illumination conditions, our method can learn a coarse-to-fine neural deformation model without considering any 3D template or the camera locations. Our algorithm retrieves RGB appearance, an implicit canonical 3D shape as well as a time-varying deformation model. Once learned, our algorithm can be also employed for novel image synthesis, where after providing a novel point of view, a photo-realistic image of the shape is inferred.

**Abstract.** We present a coarse-to-fine neural deformation model to simultaneously recover the camera pose and the 4D reconstruction of an unknown object from multiple RGB sequences in the wild. To that end, our approach does not consider any pre-built 3D template nor 3D training data as well as controlled illumination conditions, and can sort out the problem in a self-supervised manner. Our model exploits canonical and image-variant spaces where both coarse and fine components are considered. We introduce a neural local quadratic model with spatio-temporal consistency to encode fine details that is combined with canonical embeddings in order to establish correspondences across sequences. We thoroughly validate the method on challenging scenarios with complex and real-world deformations, providing both quantitative and qualitative evaluations, an ablation study and a comparison with respect to competing approaches. Our project is available at https://github.com/smontode24/4DPV.

## 1 Introduction

The problem of jointly estimating rigid 3D shape and camera pose from RGB images, known as Structure-from-Motion (SfM), has seen great progress [1, 28, 32] in the last two decades. These methods have shown wide robustness to infer accurate 3D models from visual correspondences in a large collection of non-controlled images.

Later, these algorithms were extended to handle non-rigid objects, coining the problem of Non-Rigid Structure from Motion (NRSfM). In contrast to the rigid counterpart, the non-rigid case is an ill-posed problem where the use of additional priors was needed [7, 13, 19, 20, 44]. In both cases, the algorithms relied on establishing visual correspondences throughout the dataset, a task that can become especially complex in non-rigid objects. Moreover, as the video sequences used to solve the non-rigid problem are short, these approaches can only retrieve the object visible part in the video, hindering the reconstruction of a full and 3D volumetric object. Some approaches have addressed this problem for rigid objects [39] and non-rigid ones such as humans [23] and animals [53, 54] by assuming large amounts of 3D training data. Unfortunately, acquiring volumetric 3D shapes for some objects, such as animals, is a hard task due to the standard 3D scanning systems used to capture human motion are not applicable to those scenarios in the wild [2]. The large diversity of animals and 3D configurations they can take [3] makes the problem extremely hard, particularly in unconstrained environments with unknown illumination conditions and non-controlled camera locations.

More recently, differentiable rendering approaches [46] were presented to solve the problem without assuming explicit correspondences between the images. NeRF [24] proposed a volume rendering approach by incorporating neural implicit functions to model both RGB appearance and rigid 3D shape. This work was later extended to the non-rigid domain [30, 31, 38], but assuming the camera locations in advance, unlike what SfM methods do. Some works [48] have relaxed that prior knowledge obtaining coarse results. Mainly that is due to the use of simple non-rigid models that cannot capture complex and detail-aware deformations.

In this paper, we propose a coarse-to-fine neural deformation model to jointly infer the camera pose and the 4D reconstruction of an unknown object in a non-controlled scenario. To this end, our algorithm exploits the visual information in multiple RGB videos without considering 2D matches or the use of 3D templates. We use a time-invariant canonical space together with a time-variant image one, where both coarse and fine components are considered. Following [46–48], we employ a Linear-Blend-Skinning (LBS) model to encode the coarse shape, and include a local quadratic deformation model to capture higher details. Thanks to our algorithm, we can capture challenging 4D animal motions in a self-supervised manner, outperforming state-of-the-art approaches. Figure 1 illustrates the overall idea behind our method. Our method is capable of capturing a wide variety of non-rigid and volumetric bodies, including animals where the amount of data is more restrictive to apply deep supervised learning.

## 2   Related work

**Non-rigid reconstruction.** The problem of jointly recovering a non-rigid 3D shape and the camera motion from a set of 2D point tracks in a monocular video is denoted as NRSfM. A wide body of work in this context uses probabilistic priors in the spatial [12, 37], trajectory [7], shape-trajectory [13] or force [5] domain. Subsequently, models based on an union of subspaces were introduced to handle complex deformations [52], scenarios with multiple shapes [4, 18], or rigid and non-rigid categories [3]. In parallel, some parametric models were inspired by physics and geometric models, including linear elasticity [6], LBS [21], or isometric [50] and quadratic models [10], just to name a few. More recently, the previous models were introduced in neural formulations [15,33,42] to infer the 3D shape given the 2D annotations at test time. Unfortunately, these methods rely on long-range correspondences that are not easy to obtain in practice. To avoid that, a new line of work has emerged where the feature correspondences can be implicitly handled. Lately, differentiable rendering, like LASR [46] or ViSER [47], was proposed to retrieve articulated shapes and motion from a monocular video. While these approaches provided promising results, the solution still included unrealistic 3D configurations and artifacts.

**Neural radiance fields.** NeRF [24] proposed a rigid scene neural representation for novel view synthesis from a set of images where the camera location is known. This approach has received great attention and has seen a fast-paced improvement [25, 29, 40,43,51]. NeRF has been also extended to non-rigid scenes by deforming the observed points to a canonical space or over time [8, 22, 30, 31, 38]. For instance, D-NeRF [31] proposed to deform points in a ray of a given time and obtain their corresponding point in a canonical scene, showing its plausibility for just synthetic scenes. Nerfies [30] proposed to add an elastic term that allows specific regions to have a non-local rigid transformation. NeRFPlayer [35] splits the scene into new, deformed and static content, which allows better modeling of newly introduced objects during a sequence. K-Planes [11] represents static and dynamic scenes with a varying number of planes which more efficiently represents the scene. However, the previous approaches were only evaluated in sequences with limited non-rigid motion, synthetic deformations and short videos. Moreover, these previous models typically estimate the camera parameters beforehand using COLMAP [32]. Therefore, camera parameters are estimated based on the rigid part of the scene, limiting its application to scenes where the background is dominant enough for applying SfM methods. BANMo [48] proposed a non-rigid NeRF algorithm where both the dynamic 3D shape and the camera parameters were simultaneously estimated, as NRSfM approaches do. To this end, a neural LBS model was exploited together with a canonical space. The neural model estimated a Signed Distance Function (SDF) which provided a principled way of extracting a mesh as the zero level-set of the SDF. For estimating camera parameters and regularizing the final surface, expected points on the surface were computed and matched to Continuous Surface Embeddings (CSE) [26]. While this approach provided striking results, the solution cannot capture fine details in the shape. Other approaches propose to learn an animatable kinematic chain for any articulated object [16].

**Our contributions.** We depart from previous work in that our solution can learn a coarse-to-fine neural deformation model without assuming any 3D template or precomputed camera locations in a self-supervised manner. To this end, given multiple RGB videos of an unknown and non-rigid shape with unknown illumination conditions, our approach models the object in canonical and image spaces by using implicit functions to represent the RGB appearance, 3D shape and dense semantic embedding of the shape. To achieve fine reconstructions, the implicit representation is expressed by a neural quadratic model with spatio-temporal consistency. For every new image, our 3D model is deformed and used to render an image for supervision, via differentiable volume rendering.

## 3 Revisiting Neural Radiance Fields

In this section, we introduce some ideas from NeRF [24] and adapt its formulation to the non-rigid domain. Let $\mathbf{x} = (x, y, z)$ be a 3D location, $\mathbf{d}(\zeta, \eta)$ a 3D viewing Cartesian unit direction with $\zeta$ azimuth and $\eta$ elevation, $\mathbf{c} = (r, g, b)$ an emitted color and $\sigma$ a volume density. The goal is to define a continuous shape by a plenoptic function as $P : (\boldsymbol{\theta}, \boldsymbol{\psi}) \to (\mathbf{c}, \sigma)$, where $\boldsymbol{\theta} = \gamma(\mathbf{x})$ and $\boldsymbol{\psi} = \gamma(\mathbf{d})$, being $\gamma : \mathbb{R} \to \mathbb{R}^{2L}$ a high-frequency-aware encoding with $L$ a scalar value. Basically, $\gamma(\cdot)$ is introduced to better approximate both positional and viewing-direction high frequencies. With these ingredients, the previous function can represent the shape as the volume density and directional emitted radiance at any 3D point. To this end, we also need to define a point along the ray $\mathbf{r}(\tau) = \mathbf{o} + \tau\mathbf{d}$ by its origin and the direction vector, being $\tau$ a scalar to encode a particular location in the ray. The volume density $\sigma(\mathbf{x})$ can be interpreted as the probability of a ray terminating at particle $\mathbf{x}$. The expected color $\boldsymbol{\kappa}(\mathbf{r})$ of a ray $\mathbf{r}(\tau)$ in a non-rigid shape can be written as:

$$\boldsymbol{\kappa}(\mathbf{r}) = \int_{\tau_n}^{\tau_f} A(\tau)\sigma(d_r(\boldsymbol{\theta}, \tau))\mathbf{c}(d_r(\boldsymbol{\theta}, \tau), \boldsymbol{\psi})d\tau, \qquad (1)$$

$$\text{with } A(\tau) = \exp\bigg(-\int_{\tau_n}^{\tau} \sigma(d_r(\boldsymbol{\theta}, s))ds\bigg),$$

where $\tau_n$ and $\tau_f$ are the near and far planes, respectively; and $A(\tau)$ is the accumulated probability of a ray ending at a certain ray point. $d_r$ incorporates time-varying positional information, i.e., the shape can deform over time. To represent time-dependent information, we will use the super-index $^t$, and consider a dataset composed of $T$ monocular images. Unfortunately, handling directly the problem in Eq. (1) is complex and additional priors are needed.

Inspired by NRSfM approaches [6, 10, 37], the time-varying shape can be represented by the combination of a shape at rest and an image-dependent deformation. Some recent works [31, 48] have exploited this idea to model the appearance of the non-rigid shape in a canonical space together with a neural deformation model to encode the image-by-image variation. Without loss of generality, non-rigid approaches just model the object of interest as it was common in NRSfM approaches [12, 33] or in non-rigid NeRF [31, 48], instead of considering rigid background in the estimation like Nerfies does [30].

### 3.1 Canonical space: Rest shape

To infer shape and appearance for an $n$-th 3D point in the canonical space $\mathbf{x}_n^c$, we can approximate the plenoptic function $P$ with a MLP network. In order not to abuse notation, the sub-index $n$ will be omitted. Particularly, both properties can be inferred by:

$$\mathbf{c}^t = \mathbf{MLP}_c(\mathbf{x}^c, \mathbf{d}^t, \boldsymbol{\omega}_c^t), \qquad (2)$$

$$\sigma = \Theta_\alpha(\mathbf{MLP}_{SDF}(\mathbf{x}^c)), \qquad (3)$$

where $\boldsymbol{\omega}_c^t$ is a learnable environment code to capture illumination conditions across images in the video collection. The network $\mathbf{MLP}_{SDF}$ is used to infer the shape by computing the SDF of a point to the surface that is then converted to density by means of the cumulative of a unimodal distribution with zero mean and $\alpha$ scale by the operator $\Theta_\alpha(\cdot)$ [49]. Additionally, a learned 16-dimensional canonical embedding is also recovered as $\boldsymbol{\varphi} = \mathbf{MLP}_\varphi(\mathbf{x}^c)$. The network maps 3D points to a canonical embedding that are matched by pixels from different visual conditions, enabling long-range correspondence across sequences [26, 27].

Following [48], we define a couple of time-dependent warping functions $\Upsilon^{c \to t}$ and $\Upsilon^{t \to c}$, to map the canonical location to the $t$-th image space one and its inverse mapping, respectively. In other words, $\Upsilon^{c \to t}$ is a forward warping function and $\Upsilon^{t \to c}$ is a backward warping one. With these ingredients, we only need to define a non-rigid model to encode the image-varying deformation.

Considering the previous transformations, we can now render images by volume rendering [24] but deforming the rays. Let $\bar{\mathbf{x}}^t$ be a pixel location at the $t$-th image and $\mathbf{x}_h^t$ the $h$-th 3D point sampled along the ray emanating from $\bar{\mathbf{x}}^t$. Both color and opacity $o \in [0,1]$ can be obtained as $\mathbf{c}(\bar{\mathbf{x}}^t) = \sum_{h=1}^H \mu_h \mathbf{c}^t(\Upsilon^{t \to c}(\mathbf{x}_h^t))$ and $o(\bar{\mathbf{x}}^t) = \sum_{h=1}^H \mu_h$, where $\mu_h = \prod_{g=1}^{h-1} \rho_g (1 - \rho_h)$ is the probability the point $\mathbf{x}_h^t$ is visible to the camera. $\rho_h = \exp(-\sigma_h \delta_h)$ represents the probability that a photon is transmitted between consecutive samples with interval $\delta_h$, and $\sigma_h = \sigma(\Upsilon^{t \to c}(\mathbf{x}_h^t))$ is the density in Eq. (3). The ray-surface intersection in the canonical space can then be computed as $\mathbf{x}^e(\bar{\mathbf{x}}^t) = \sum_{h=1}^H \mu_h(\Upsilon^{t \to c}(\mathbf{x}_h^t))$.

### 3.2 Image space: Deformable shape

To encode the deformation, recent works [48] have used a neural blend-skinning model [17, 21]. Let $\mathbf{x}^t$ be a 3D point in the $t$-th image space that can be transformed to $\mathbf{x}^c$ by blending the rigid transformations of $B$ bones. To that end, a transformation of the shape root body from canonical space to $t$-th image space is required as $\mathbf{G}^t \in SE(3)$ as well as a rigid transformation that deforms the $b$-th bone from its canonical state to the deformed one by means of $\mathbf{J}_b^t \in SE(3)$. Both root pose $\mathbf{G}^t$ and body pose $\mathbf{J}_b^t$ with 3D translations and rotations can be inferred by:

$$\mathbf{G}^t = \mathbf{MLP}_G(\boldsymbol{\omega}_r^t), \tag{4}$$

$$\mathbf{J}_b^t = \mathbf{MLP}_J(\boldsymbol{\omega}_b^t), \tag{5}$$

where $\boldsymbol{\omega}_r^t$ and $\boldsymbol{\omega}_b^t$ are 128-dimensional latent codes for root pose as well as body pose at frame $t$, respectively. For camera parameters, we use an initial root pose $\mathbf{G}_0^t$ by applying PoseNet [26], and then the root pose can be refined as $\mathbf{G}^t = \mathbf{MLP}_G(\boldsymbol{\omega}_r^t)\mathbf{G}_0^t$. Both mappings can be defined as:

$$\mathbf{x}^t = \mathbf{G}^t((\sum_{b=1}^B \mathbf{W}_b^c \mathbf{J}_b^t)\mathbf{x}^c) = \Upsilon^{c \to t}(\mathbf{x}^c), \tag{6}$$

$$\mathbf{x}^c = (\sum_{b=1}^B \mathbf{W}_b^t (\mathbf{J}_b^t)^{-1})((\mathbf{G}^t)^{-1}\mathbf{x}^t) = \Upsilon^{t \to c}(\mathbf{x}^t), \tag{7}$$

where $\mathbf{W}_b^c$ and $\mathbf{W}_b^t$ are the blend skinning weights for points $\mathbf{x}^c$ and $\mathbf{x}^t$ relative to the $b$-th bone.

Each bone is encoded by a bone center $\mathbf{e}_b$, bone orientation and bone scale. The last two are considered with a precision matrix $\mathbf{Q}_b$. Therefore, bones are represented with 3D Gaussian ellipsoids that are centered at $\mathbf{e}_b$. To determine the attachment of a vertex $\mathbf{x}$ to each bone $b$, skinning weights for coarse geometry are computed by a Mahalanobis distance [46] and the ones for fine one by a MLP [48] as $\mathbf{W}_b = \iota((\mathbf{x} - \mathbf{e}_b)^\top \mathbf{Q}_b (\mathbf{x} - \mathbf{e}_b) + \mathbf{MLP}_W(\mathbf{x}, \boldsymbol{\omega}_b))$, where $\mathbf{Q}_b = \mathbf{V}_b^\top \boldsymbol{\Xi}_b^0 \mathbf{V}_b$ is composed of bone's orientation $\mathbf{V}_b$ and diagonal scaling matrices $\boldsymbol{\Xi}_b^0$, that represent their size. $\iota$ indicates a softmax normalization operator. To infer $\mathbf{W}_b^c$ and $\mathbf{W}_b^t$, we apply the previous expression to the canonical or $t$-th space points, respectively.

## 4  Our Approach

Thanks to the LBS model that was considered in Section 3.2, promising results can be obtained [48]. However, it is worth noting that the LBS model was initially proposed for articulated bodies [17, 21] and therefore, it is just effective in modeling the coarse representation of the continuous shape, being unable to encode the deformations of a detail-aware shape. To address that limitation, in this paper we propose to combine the LBS model to capture coarse deformations with a local quadratic deformation model inspired by [9, 10] to capture the fine ones. To this end, we introduce a fine deformation MLP network that given the coarse canonical location denoted by $\mathbf{x}_c^c$ (see Section 3.2) and a 128-dimensional deformation code $\mathbf{h}^t$, it recovers a deformation to produce the fine estimation $\mathbf{x}_f^c$. $\mathbf{h}^t$ is jointly optimized with gradient descent like the rest of model parameters.

### 4.1  Fine Image space: Quadratic Deformation

We propose a local neural quadratic deformation model (LQM) that can consider both linear and quadratic behaviors as well as incorporate cross-term components to increase the variety of deformations. Our model is available to capture bending, stretching and twisting deformations with a simple but effective formulation. We first define a $9 \times n$ matrix of extended coordinates $\mathbf{D}(\mathbf{X}_c^c)$ that contains the 3D coordinates (and its quadratic and cross-term variants) of the $N$ points in the canonical coarse space:

$$\mathbf{D} = \begin{bmatrix} x_1 & y_1 & z_1 & x_1^2 & y_1^2 & z_1^2 & x_1 y_1 & y_1 z_1 & z_1 x_1 \\ \vdots & \vdots & \vdots & \vdots & \vdots & \vdots & \vdots & \vdots & \vdots \\ x_n & y_n & z_n & x_n^2 & y_n^2 & z_n^2 & x_n y_n & y_n z_n & z_n x_n \end{bmatrix}^\top. \quad (8)$$

In order to infer the deformation $\mathbf{X}^t$ at image $t$ of the $N$ points from the canonical coarse configuration $\mathbf{X}_c^c$, we define the quadratic transformation:

$$\mathbf{X}^t = \begin{bmatrix} \boldsymbol{\Gamma}^t & \boldsymbol{\Omega}^t & \boldsymbol{\Lambda}^t \end{bmatrix} \mathbf{D}(\mathbf{X}_c^c) = \mathbf{A}^t \mathbf{D}(\mathbf{X}_c^c), \quad (9)$$

where $\boldsymbol{\Gamma}^t$, $\boldsymbol{\Omega}^t$, and $\boldsymbol{\Lambda}^t$ are $3 \times 3$ transformation matrices associated with the linear, quadratic and cross-term coefficients of the shape at image $t$, respectively. For simplicity, we also define a $3 \times 9$ matrix $\mathbf{A}^t$ to collect all the weight coefficients in the deformation model. Note that we may apply the same $\mathbf{A}^t$ for all the points in the shape, obtaining global deformations that cannot include local and complex ones. To avoid that, we define $\mathbf{A}_n^t$ at point $n$, which is a local version of $\mathbf{A}^t$.

The weight coefficients in $\mathbf{A}_n^t$ are inferred with a MLP network, and they can then be used to compute the fine deformed point at the canonical frame and $n$-th point as:

$$\mathbf{x}_{f,n}^c = \mathbf{A}_n^t \mathbf{D}(\mathbf{x}_{c,n}^c), \tag{10}$$

where $\mathbf{x}_{c,n}^c$ is the $n$-th coarsely deformed point in the canonical space. Both terms in Eq. (10) can be obtained as:

$$\mathbf{A}_n^t = \mathbf{MLP}_Q(\Upsilon^{t \to c}(\mathbf{x}_n^t), \mathbf{h}^t) = \mathbf{MLP}_Q(\mathbf{x}_{c,n}^c, \mathbf{h}^t), \ \mathbf{D}(\mathbf{x}_{c,n}^c) = \mathbf{D}(\Upsilon^{t \to c}(\mathbf{x}_n^t)), \tag{11}$$

where the weight coefficients are given by $\mathbf{MLP}_Q$.

To increase robustness in our local and quadratic deformation model, we incorporate a spatio-temporal local coherence in the deformation. To this end, we exploit a neighborhood of $K$ points to enforce a similar deformation. First, we propose a loss function to impose that constraint in the spatial domain as:

$$\mathcal{L}_Q^s = \sum_{n=1}^{N} \sum_{k=1}^{K} ||\mathbf{A}_n^t - \mathbf{A}_{s,k}^t||_{\mathcal{F}} \tag{12}$$

$$= \sum_{n=1}^{N} \sum_{k=1}^{K} ||\mathbf{MLP}_Q(\mathbf{x}_{c,n}^c, \mathbf{h}^t) - \mathbf{MLP}_Q(\mathbf{x}_{c,n}^c + \mathbf{x}_{s,k}^t, \mathbf{h}^t)||_{\mathcal{F}},$$

where $\mathbf{x}_{s,k}^t \sim \mathcal{N}(\mathbf{0}, \epsilon \mathbf{I}_3)$, with $\mathbf{I}_3$ a $3 \times 3$ identity matrix. $||\cdot||_{\mathcal{F}}$ represents a Frobenius norm of a matrix. Second, we also propose a loss function to impose the constraint in the temporal domain as:

$$\mathcal{L}_Q^t = \sum_{n=1}^{N} \sum_{k=1}^{K} ||\mathbf{A}_n^t - \mathbf{A}_{s,k}^{t+1}||_{\mathcal{F}} \tag{13}$$

$$= \sum_{n=1}^{N} \sum_{k=1}^{K} ||\mathbf{MLP}_Q(\mathbf{x}_{c,n}^c, \mathbf{h}^t) - \mathbf{MLP}_Q(\mathbf{x}_{c,n}^c + \mathbf{x}_{s,k}^{t+1}, \mathbf{h}^{t+1})||_{\mathcal{F}},$$

where $\mathbf{x}_{s,k}^{t+1} \sim \mathcal{N}(\mathbf{0}, \epsilon \mathbf{I}_3)$. Then, the total loss is $\mathcal{L}_Q = \mathcal{L}_Q^s + \lambda_{qt} \mathcal{L}_Q^t$, with $\lambda_{qt}$ a weight factor.

While the coarse network in Section 3.2 optimizes a coarse shape in the $t$-th image space together with the camera pose, we exploit the fine network for capturing as many details as possible. Thanks to this new detail-aware deformation, our global algorithm can recover more accurate and realistic reconstructions than competing approaches.

### 4.2 Fine Canonical Space

To finish, we introduce a fine canonical-space network to recover even more accurate estimations. We have observed that the CSE-matching of expected 3D points in the shape along with incorrect optical-flow estimations over-constrain the neural-radiance-field algorithm in such a way that the network is only able to model coarse representations. As a consequence, correct registration of finer details cannot be achieved. To address that, we propose to incorporate a fine network $\mathbf{MLP}_f$ that has the same architecture as the coarse one in Section 3.1, i.e., Eqs. (2)-(3), but considering additional samples to render a particular pixel.

## 5  Optimization

The model is learned by minimizing a differentiable objective function which is optimized using stochastic gradient descent in a self-supervised fashion. To this end, we next define a loss function that includes photometric, silhouette and optical-flow terms as well as some additional regularizations. Following differentiable rendering approaches [24], we exploit a photometric loss in the coarse $\mathcal{L}_{pho}^c$ and fine $\mathcal{L}_{pho}^f$ states as:

$$\mathcal{L}_{pho}^c = \sum_{\bar{\mathbf{x}}} \|\mathbf{c}_c(\bar{\mathbf{x}}) - \mathbf{c}(\bar{\mathbf{x}})\|_1, \ \mathcal{L}_{pho}^f = \sum_{\bar{\mathbf{x}}} \|\mathbf{c}_f(\bar{\mathbf{x}}) - \mathbf{c}(\bar{\mathbf{x}})\|_1, \tag{14}$$

where $\mathbf{c}_c$ and $\mathbf{c}_f$ are the rendered color by the coarse and fine networks, respectively; and $\mathbf{c}$ the corresponding ground truth for a particular pixel location $\bar{\mathbf{x}}$. We choose to use a $l_1$-norm in the photometric loss as it was experimentally observed that gives the best results. Moreover, a silhouette loss is also included to penalize the deviations between the predicted 2D shape in the image and the corresponding segmentation mask. Like in the photometric loss, this term is applied for both coarse and fine networks as:

$$\mathcal{L}_{sil}^c = \sum_{\bar{\mathbf{x}}} \|\mathbf{s}_c(\bar{\mathbf{x}}) - \mathbf{s}(\bar{\mathbf{x}})\|_2^2, \ \mathcal{L}_{sil}^f = \sum_{\bar{\mathbf{x}}} \|\mathbf{s}_c(\bar{\mathbf{x}}) - \mathbf{s}_f(\bar{\mathbf{x}})\|_2^2, \tag{15}$$

where $\mathbf{s}(\bar{\mathbf{x}})$ denotes the segmentation mask of the object in the image containing the pixel location $\bar{\mathbf{x}}$; $\mathbf{s}_c(\bar{\mathbf{x}})$ and $\mathbf{s}_f(\bar{\mathbf{x}})$ refer to the silhouette prediction of the coarse and fine networks, respectively. The fine network takes advantage of the learned knowledge of the coarse one which accurately represents the shape and tries to predict the same silhouette as the coarse network, acting as a teacher to the fine network. Although the coarse network recovers the low frequencies and the fine the high ones, the silhouette in the image is quite similar.

Optical flow is also considered to constrain the solution, by penalizing the difference between the rendered 2D flow and the estimation by an off-the-shelf flow network. The expected surface point is computed in the canonical space from deformed points on the ray $\mathbf{r}$ at the $t$-th image of pixel $\bar{\mathbf{x}}$. This point is then deformed in the $t+1$-th image, where it is reprojected through the camera matrix $\mathbf{P}^{t+1}$ as:

$$\mathcal{L}_{OF} = \sum_{\bar{\mathbf{x}}^t} \|\mathbf{P}^{t+1} \Upsilon^{c \to t+1}(\mathbf{x}^c(\bar{\mathbf{x}}^t)) - \mathbf{v}^{t+1}(\bar{\mathbf{x}}^t)\|_2^2, \tag{16}$$

where $\mathbf{v}^{t+1}(\bar{\mathbf{x}}^t)$ denotes the optical flow measured between frames $t$ and $t+1$ in the pixel $\bar{\mathbf{x}}$. $\mathbf{P}^t$ represents the projection matrix of a pinhole camera at the $t$-th image. This loss will only be considered in the coarse network, as it was discussed in Section 4.2.

We now include some additional regularization losses to better infer the shape. While these terms can help to recover the global 3D configuration, they may limit the capture of fine details that will be acquired thanks to our fine deformation model. We define a couple of canonical embedding matching losses in 3D and 2D, respectively. In the first case, an expected 3D point $\mathbf{x}^c(\mathbf{x}^t)$ is matched to the closest interpolated point in the canonical surface embedding $\tilde{\mathbf{x}}^c(\mathbf{x}^t)$. In the second one, the rendered CSE in the image should match those predicted by a DensePose network [26], denoted as $D_P(\cdot)$. Both losses can be written as:

$$\mathcal{L}_{cse3D} = \sum_{\mathbf{x}^t}||\mathbf{x}^c(\mathbf{x}^t) - \tilde{\mathbf{x}}^c(\mathbf{x}^t)||_2^2, \; \mathcal{L}_{cse2D} = \sum_{\bar{\mathbf{x}}^t}||\mathbf{P}^t\mathbf{x}^c(\bar{\mathbf{x}}^t) - \mathbf{P}^t D_P(\bar{\mathbf{x}}^t)||_2^2. \quad (17)$$

Similar to [17, 22, 48], we apply both 2D and 3D cycle consistencies. To that end, the expected surface point on a ray should be reprojected into the pixel inside the ray. In the same way, a forward mapping followed by a backward mapping to a 3D point should be mapped into itself. The previous ideas can be imposed by the loss terms:

$$\mathcal{L}_{cyc2D} = \sum_{\bar{\mathbf{x}}^t}||\mathbf{P}^t\Upsilon^{c\to t}(\mathbf{x}^c(\bar{\mathbf{x}}^t)) - \bar{\mathbf{x}}^t||_2^2, \; \mathcal{L}_{cyc3D} = \sum_{h}\mu_h||\Upsilon^{c\to t}(\Upsilon^{t\to c}(\mathbf{x}_h^t)) - \mathbf{x}_h^t||_2^2.$$

Finally, we include a temporal smoothness term in the camera motion by employing a first-order filter as:

$$\mathcal{L}_{cam} = \sum_t ||\bar{\mathbf{P}}^t - \bar{\mathbf{P}}^{t+1}||_2^2, \quad (18)$$

where $\bar{\mathbf{P}}^t$ denotes the $t$-th extrinsic parameters of the camera, i.e., including both rotation and translation values.

The global objective function to minimize is the sum of all losses as:

$$\mathcal{L} = \underbrace{(\lambda_{pf}\mathcal{L}_{pho}^f + \lambda_{sf}\mathcal{L}_{sil}^f + \lambda_Q\mathcal{L}_Q)}_{\text{Fine NeRF network}} + \underbrace{(\lambda_{pc}\mathcal{L}_{pho}^c + \lambda_{sc}\mathcal{L}_{sil}^c + \lambda_{of}\mathcal{L}_{OF})}_{\text{Coarse NeRF network}}$$
$$+ \underbrace{\lambda_{reg}(\mathcal{L}_{cse3D}^c + \mathcal{L}_{cse2D}^c + \mathcal{L}_{cse2D}^f)}_{\text{CSE Regularization}} + \underbrace{\mathcal{L}_{cyc2D} + \mathcal{L}_{cyc3D} + \mathcal{L}_{cam}}_{\text{Other regularization}}, \quad (19)$$

where $\lambda_{pf}$, $\lambda_{sf}$, $\lambda_Q$, $\lambda_{pc}$, $\lambda_{sc}$, $\lambda_{of}$ and $\lambda_{reg}$ are hyperparameters weighting each loss term. All MLPs weights, learnable latent codes and camera parameters are jointly optimized by minimizing $\mathcal{L}$. Figure 2 illustrates the overall method we propose.

## 6 Experimental Evaluation

In this section, our experimental results for several types of scenarios are presented, including synthetic and real datasets (see videos of the experimental results in the supplementary material). We provide both qualitative and quantitative evaluations as well as comparisons w.r.t. competing approaches. For all the synthetic experiments, we provide two metrics: 1) a mean 3D Chamfer Distance (CD) error to quantify the quality of the reconstruction between the estimated and ground-truth mesh, and 2) the F-score [36] at distance thresholds that are set to the 2% of the longest edge (denoted as F@2%).

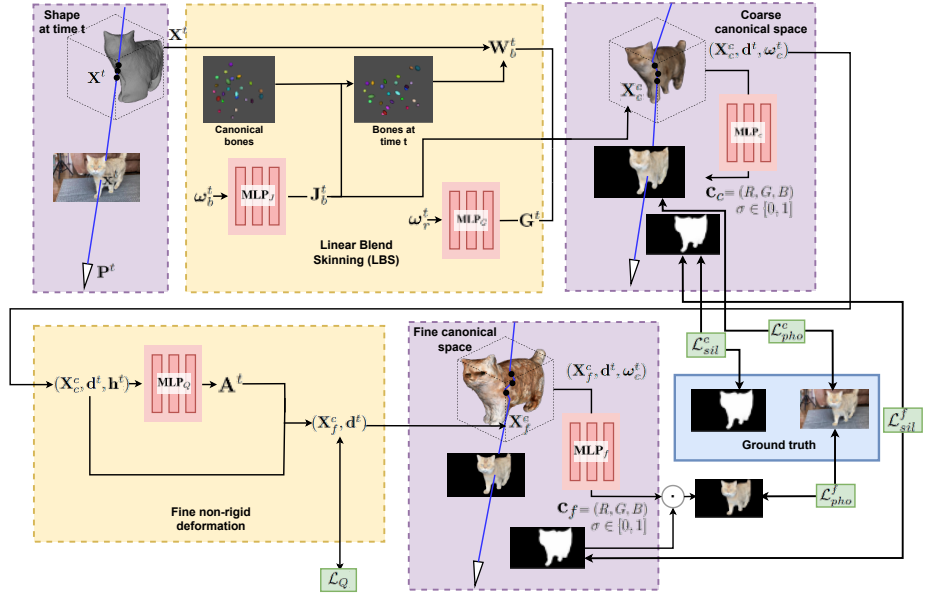

**Fig. 2: Self-supervised coarse-to-fine approach to capture 4D shapes from multiple videos.** LBS model points at image $t$ are mapped into the coarse canonical space. Then, they are refined with the fine regularized non-rigid deformation and an additional fine canonical network to render the RGB image.

### 6.1 Optimization and Implementation details

Our implementation of implicit shape and appearance models uses NeRF [24] and BANMo [48]. The coarse and fine MLPs are each made of $8$ layers with hidden dimensionality of $256$ using ReLU as a non-linear activation. The fine deformation $\mathbf{MLP}_Q$ is made of $6$ layers with hidden dimensionality of $128$. Similar to BANMo [48], we set the number of bones to $25$ and extract the mesh running marching cubes to extract the zero level-set of the SDF. The marching cubes resolution grid is set to $512^3$ for general analysis and to $1024^3$ for detailed comparison. The SDF is extracted from the fine network. The number of sampled neighbors $K$ in the deformation smoothness constraints is set to $6$. The combination of hyperparameters leading to the best result are $\lambda_{pf} = 0.1$, $\lambda_{sf} = 1$, $\lambda_Q = 1e3$, $\lambda_{pc} = 0.1$, $\lambda_{sc} = 1$, $\lambda_{of} = 0.1$ and $\lambda_{reg} = 0.02$. Every experiment is run for 224k iterations using an Adam optimizer. We obtain the weight coefficients experimentally and fix them for all experiments, obtaining a non-overfitted solution.

We follow a multistep optimization approach composed of two phases. **First phase:** A warmup phase of 9,600 iterations is performed where the optical flow loss $\mathcal{L}_{OF}$ is given more weight and is linearly decayed with the same setup as BANMo [48]. Once the warmup stage finishes, bones are reinitialized for better convergence and more robust minimization as done in previous models using the LBS model [46, 48]. After the warmup, we train it for 14,400 more iterations. Fine samples and active ones (samples with large photometric error) are not used in this phase. **Second phase:** The bones and learning rate are reinitialized and all losses are enabled. Active samples are used for addressing the most difficult regions and fine samples are allocated in high-density regions. **MLP**$_G$'s weights and $\omega_r^t$ are no longer updated, therefore camera parameters remain the same in this phase for better stabilization and to ease the convergence. We sample 256 coarse samples in the coarse network and half of them are sampled in the fine one as important samples. The number of frequency bands for the position and viewing-direction encoding in Section 3 is $L = 10$ and $L = 4$, respectively. We use a batch size of 416 rays and optimize each model with PyTorch on a single NVIDIA RTX 3090 GPU for about one day. This training time corresponds to training on datasets consisting of approximately 2,000 frames; however, the time does not increase significantly for larger sequences.

### 6.2 Datasets

As it was commented above, we propose to use both synthetic and real videos for evaluation. For quantitative evaluation, we select four datasets from [48]. Particularly, we use 2,600 frames of two sets of videos of the AMA human dataset [41] that contain multi-view videos captured by eight synchronized cameras, where a high-fidelity ground truth is given. These datasets will be denoted by *AMA-swing* and *AMA-samba*. We also consider the animated 3D datasets *eagle* and *hands*. For every scenario, we have five videos with 150 frames per video with known root poses and silhouettes.

Note that evaluating the dense 4D reconstruction from multiple videos is hard as the capture of a ground truth geometry and radiance properties of the shape is not trivial in practice. As a consequence, we propose to use three real datasets for quantitative evaluation, where no ground truth is available. In all cases, the non-rigid object was recorded in the wild by a phone camera in non-controlled conditions. That means, in multiple videos, both the object and the camera can freely move in the 3D space, while the object can exhibit large and complex non-rigid deformations, as well as different illumination conditions are also unknown. In the first case, we consider the *casual-cat* dataset proposed in [48] composed of 11 videos and 1,973 images in total. In the second one, we use 10 videos with 583 images in total of *adult-5* dataset, where a human is performing some motions. In the third one, we record a casual video in the wild of a dog, denoted as *casual-dog*, while it moves, deforms, and interacts with the environment. The dataset is composed of 6 videos and 4,071 images in total. The preprocessing pipeline consists of extracting the object silhouette and optical flow with off-the-shelf models, such as PointRend [14] and VCN-robust [45].

### 6.3 Reconstruction and comparison

First of all, we assess the importance of each component in our approach by quantitatively and qualitatively evaluating our results and comparing them with competing approaches. For comparison, we consider the results reported in BANMo [48] and ViSER [47]. Alternatively, we also employ this algorithm in our training setup, a more environment-aware scenario with limited computational resources (as it was commented in Section 6.1) that we consider our baseline. With this starting point, we present two intermediate solutions (just adding the fine canonical network and just adding the fine deformation one) and the one with our full algorithm. Table 1 summarizes our results for all the synthetic datasets. As it can be seen, and even in these sequences where meshes are not very highly detailed, our solution obtains the best estimation on average, by outperforming both the original BANMo [48] and ViSER [47] solutions. Our model produces worse results in the *hands* dataset which is a synthetic coarse mesh and whose deformation can be modeled in a quasi-linear fashion. The impact of adding the fine deformation model as well as the full algorithm can be seen in Fig. 3 for the *casual-cat* dataset. We can observe that adding the fine network without 3D CSE constraining as well as coarse and fine deformation provides better modeling than competing solutions in terms of both appearance and shape (some artifacts as the extra leg are not physically possible). In Fig. 4, we provide an example of some shape deformations along with their CSE. Moreover, in Table 2, we perform an ablation study in the *casual-cat* dataset to validate the components of our approach.

| Method | AMA-swing | | AMA-samba | | Eagle | | Hands | |
|---|---|---|---|---|---|---|---|---|
| | CD ↓ | F@2% ↑ | CD ↓ | F@2% ↑ | CD ↓ | F@2% ↑ | CD ↓ | F@2% ↑ |
| BANMo [48] | 9.1 | 57.0 | 10.6 | - | 8.1 | 56.7 | 7.5 | 49.6 |
| ViSER [47] | 15.7 | 52.2 | - | - | 23.0 | 20.6 | 16.8 | 21.3 |
| Baseline | 10.9 | 41.1 | **9.2** | 55.5 | 7.4 | 67.2 | 5.4 | 57.4 |
| + fcn | 10.6 | 45.7 | 9.8 | 57.1 | **4.7** | **81.2** | **4.8** | **73.5** |
| + fdn | 10.9 | 41.1 | 11.7 | 49.4 | 6.2 | 65.3 | 6.8 | 54.4 |
| Ours | **10.0** | **50.3** | 9.4 | **61.1** | 5.1 | 74.5 | 6.0 | 56.2 |

**Table 1: Quantitative comparison on synthetic datasets.** The table reports the average Chamfer distance (CD) in cm and the F@2% for two training cases. In the first case, the solution in BANMo [48] and ViSER [47] are considered. In the second case, a more environment-aware training process with limited resources is proposed, including the solution after directly running BANMo [48] (this version is denoted by Baseline), and our result for: just adding the fine canonical network (denoted by fcn) or the fine deformation one (denoted by fdn), and our full algorithm.

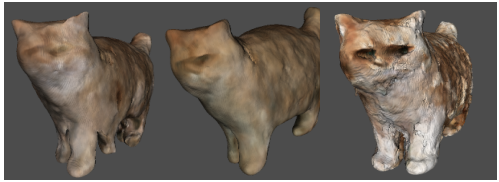

**Fig. 3: Qualitative ablation in the** *casual-cat* **dataset. From left to right**: BANMo baseline [48]; Our method with a coarse canonical network together with the fine deformation model; Our full approach with fine canonical and deformation networks.

Finally, we show qualitative results for 3D mesh extraction in Fig. 5 for the datasets *casual-cat*, *adult-5*, *eagle* and *casual-dog*. As it can be observed, our method recovers much more details in the *casual-cat* and *casual-dog* datasets in comparison with those recovered by [48], while producing more physically-aware shapes. It is worth mention-ing that, in the *eagle* dataset, our method does not produce artifacts like in [48] and better models its head and claws (see the extra head). Similar results are obtained in the *adult-5* dataset. On balance, our method works much better on the real datasets than on the synthetic ones compared to [48], since these datasets include more complex deformations that cannot be captured with piecewise linear models as done in the literature. Once our model is learned, we can also provide novel images from an alternative point of view for a particular $t$-th pose (some examples are displayed in Fig. 6).

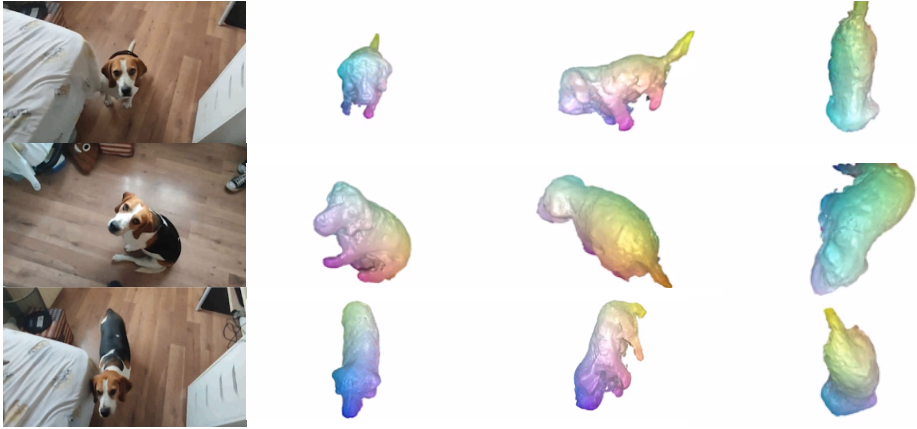

**Fig. 4:** CSE embeddings of deformed views in the *casual-dog* dataset. **From left to right**: Input frame, CSE-colored mesh visualized in camera, side and top views, respectively.

|  | Mean PSNR ↑ | Mean SSIM ↑ |
|---|---|---|
| Baseline | 33.439 | 0.957 |
| Coarse + deformation code | 35.185 | 0.954 |
| Coarse + ray smoothness | 35.418 | 0.959 |
| Coarse + quadratic deformation | 35.074 | 0.964 |
| Coarse + fine + ray smoothness | 34.658 | 0.960 |
| Coarse + fine + spatial smoothness | 33.893 | 0.958 |
| Coarse + fine + spatial + temporal smoothness | 35.231 | 0.963 |
| **Coarse + fine + LQM + smoothness (Ours)** | **35.853** | **0.967** |

**Table 2: View synthesis ablation study.** The table reports the view synthesis quality of frames in the *casual-cat* dataset.

## 7 Conclusion

In this work, we have presented a coarse-to-fine deformation method that better models and reconstructs objects from a collection of RGB videos in a self-supervised manner. Objects can deform freely, move and interact with the scenario while being observed by the camera with an unknown trajectory and under non-controlled lighting conditions. We have shown the importance of not over-constraining the network that finely recovers the shape and how a coarse deformation model can be combined with a fine one. Additionally, we have performed an in-depth experimental analysis in challenging videos in the wild, including some comparisons and an ablation study where several fine deformation models have been evaluated. Our algorithm has outperformed both quantitatively and qualitatively the solutions provided by state-of-the-art approaches. An interesting avenue for future research is to extend our algorithm to include more sophisticated hair models [34].

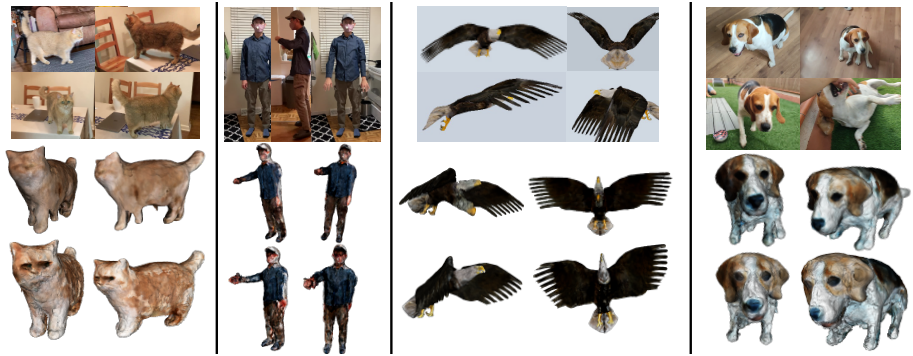

**Fig. 5: Qualitative evaluation.** The same information is provided in every column for the datasets *casual-cat*, *adult-5*, *eagle* and *casual-dog*. **Top:** Some pictures in the input dataset. **Middle:** 3D color mesh inferred by BANMo [48]. **Bottom:** Our estimation. In all cases, meshes are extracted with marching cubes at a resolution grid of $1024^3$.

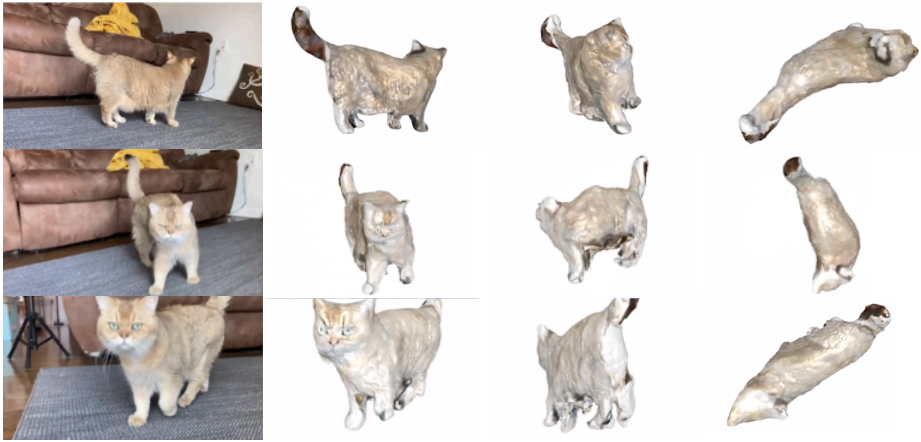

**Fig. 6: Qualitative evaluation of deformed views in the *casual-cat* dataset. From left to right:** Input frame, colored mesh visualized in camera, side and top views, respectively.

**Acknowledgment.** Work produced with the support of a 2023 Leonardo Grant for Scientific Research and Cultural Creation, BBVA Foundation.

# References

1. Agarwal, S., Snavely, N., Simon, I., Seitz, S.M., Szeliski, R.: Building Rome in a day. In: ICCV (2009)
2. Agudo, A.: Safari from visual signals: Recovering volumetric 3D shapes. In: ICASSP (2022)
3. Agudo, A.: Unsupervised 3D reconstruction and grouping of rigid and non-rigid categories. TPAMI **44**(1), 519–532 (2022)
4. Agudo, A., Moreno-Noguer, F.: DUST: Dual union of spatio-temporal subspaces for monocular multiple object 3D reconstruction. In: CVPR (2017)
5. Agudo, A., Moreno-Noguer, F.: Force-based representation for non-rigid shape and elastic model estimation. TPAMI **40**(9), 2137–2150 (2018)
6. Agudo, A., Moreno-Noguer, F., Calvo, B., Montiel, J.M.M.: Sequential non-rigid structure from motion using physical priors. TPAMI **38**(5), 979–994 (2016)
7. Akhter, I., Sheikh, Y., Khan, S., Kanade, T.: Trajectory space: A dual representation for nonrigid structure from motion. TPAMI **33**(7), 1442–1456 (2011)
8. Chen, X., Zheng, Y., Black, M.J., Hilliges, O., Geiger, A.: Snarf: Differentiable forward skinning for animating non-rigid neural implicit shapes. In: ICCV (2021)
9. Fayad, J., Agapito, L., Del Bue, A.: Piecewise quadratic reconstruction of non-rigid surfaces from monocular sequences. In: ECCV (2010)
10. Fayad, J., Del Bue, A., Agapito, L., Aguiar, P.: Non-rigid structure from motion using quadratic deformation models. In: BMVC (2009)
11. Fridovich-Keil, S., Meanti, G., Warburg, F.R., Recht, B., Kanazawa, A.: K-planes: Explicit radiance fields in space, time, and appearance. In: Proceedings of the IEEE/CVF Conference on Computer Vision and Pattern Recognition. pp. 12479–12488 (2023)
12. Garg, R., Roussos, A., Agapito, L.: Dense variational reconstruction of non-rigid surfaces from monocular video. In: CVPR (2013)
13. Gotardo, P.F.U., Martinez, A.M.: Computing smooth time-trajectories for camera and deformable shape in structure from motion with occlusion. TPAMI **33**(10), 2051–2065 (2011)
14. Kirillov, A., Wu, Y., He, K., Girshick, R.: Pointrend: Image segmentation as rendering. In: CVPR (2020)
15. Kong, C., Lucey, S.: Deep non-rigid structure from motion. In: ICCV (2019)
16. Kuai, T., Karthikeyan, A., Kant, Y., Mirzaei, A., Gilitschenski, I.: CAMM: Building category-agnostic and animatable 3D models from monocular videos. In: CVPR (2023)
17. Kulkarni, N., Gupta, A., Fouhey, D., Tulsiani, S.: Articulation-aware canonical surface mapping. In: CVPR (2020)
18. Kumar, S., Dai, Y., Li, H.: Spatio-temporal union of subspaces for multi-body non-rigid structure-from-motion. PR **77**(11), 428–443 (2017)
19. Lee, M., Cho, J., Oh, S.: Consensus of non-rigid reconstructions. In: CVPR (2016)
20. Lee, M., Choi, C.H., Oh, S.: A procrustean markov process for non-rigid structure recovery. In: CVPR (2014)
21. Lewis, J., Cordner, M., Fong, N.: Pose space deformation: a unified approach to shape interpolation and skeleton-driven deformation. In: ACM SIGGRAPH (2000)
22. Li, Z., Nuklaus, S., Snavely, N., Wang, O.: Neural scene flow fields for space-time view synthesis of dynamic scenes. In: CVPR (2021)
23. Loper, M., Mahmood, N., Romero, J., Pons-Moll, G., Black, M.J.: SMPL: A skinned multi-person linear model. TOG **34**(6), 1–16 (2015)
24. Mildenhall, B., Srinivasan, P., Tancik, M., Barron, J., Ramamoorthi, R., Ng, R.: Nerf: Representing scenes as neural radiance fields for view synthesis. In: ECCV (2020)
25. Müller, T., Evans, A., Schied, C., Keller, A.: Instant neural graphics primitives with a multiresolution hash encoding. arXiv preprint arXiv:2201.05989 (2022)

26. Neverova, N., Novotny, D., Szafraniec, M., Khalidov, V., Labatut, P., Vedaldi, A.: Continuous surface embeddings. NeurIPS (2020)
27. Neverova, N., Sanakoyeu, A., Labatut, P., Novotny, D., Vedaldi, A.: Discovering relationships between object categories via universal canonical maps. In: CVPR (2021)
28. Newcome, R., Davison, A.J.: Live dense reconstruction with a single moving camera. In: CVPR (2010)
29. Niemeyer, M., Barron, J., Mildenhall, B., Sajjadi, M., Geiger, A., Radwan, N.: Regnerf: Regularizing neural radiance fields for view synthesis from sparse inputs. In: CVPR (2022)
30. Park, K., Sinha, U., Barron, J., Bouaziz, S., Goldman, D., Seitz, S., Martin-Brualla, R.: Nerfies: Deformable neural radiance fields. In: ICCV (2021)
31. Pumarola, A., Corona, E., Pons-Moll, G., Moreno-Noguer, F.: D-nerf: Neural radiance fields for dynamic scenes. In: CVPR (2021)
32. Schonberger, J., Frahm, J.: Structure-from-motion revisited. In: CVPR (2016)
33. Sidhu, V., Tretschk, E., Golyanik, V., Agudo, A., Theobalt, C.: Neural dense non-rigid structure from motion with latent space constraints. In: ECCV (2020)
34. Sklyarova, V., Chelishev, J., Dogaru, A., Medvedev, I., Lempitsky, V., Zakharov, E.: Neural haircut: Prior-guided strand-based hair reconstruction. arXiv preprint arXiv:2306.05872 (2023)
35. Song, L., Chen, A., Li, Z., Chen, Z., Chen, L., Yuan, J., Xu, Y., Geiger, A.: Nerfplayer: A streamable dynamic scene representation with decomposed neural radiance fields. IEEE Transactions on Visualization and Computer Graphics **29**(5), 2732–2742 (2023)
36. Tatarchenko, M., Richter, S., Stephan, R., Ranftl, R., Li, Z., Koltun, V., Brox, T.: What do single-view 3D reconstruction networks learn? In: CVPR (2019)
37. Torresani, L., Hertzmann, A., Bregler, C.: Nonrigid structure-from-motion: estimating shape and motion with hierarchical priors. TPAMI **30**(5), 878–892 (2008)
38. Tretschk, E., Tewari, A., Golyanik, V., Zollhöfer, M., Lassner, C., Theobalt, C.: Non-rigid neural radiance fields: Reconstruction and novel view synthesis of a dynamic scene from monocular video. In: ICCV (2021)
39. Ulusoy, A.O., Black, M.J., Geiger, A.: Patches, planes and probabilities: A non-local prior for volumetric 3D reconstruction. In: CVPR (2016)
40. Verbin, D., Hedman, P., Mildenhall, B., Zickler, T., Barron, J., Srinivasan, P.: Ref-nerf: Structured view-dependent appearance for neural radiance fields. arXiv preprint arXiv:2112.03907 (2021)
41. Vlasic, D., Baran, I., Matusik, W., Popović, J.: Articulated mesh animation from multi-view silhouettes. In: ACM SIGGRAPH (2008)
42. Wang, C., Lucey, S.: Paul: Procrustean autoencoder for unsupervised lifting. In: CVPR (2021)
43. Xu, Q., Xu, Z., Philip, J., Bi, S., Shu, Z., Sunkavalli, K., Neumann, U.: Point-nerf: Point-based neural radiance fields. In: CVPR (2022)
44. Xu, X., Dunn, E.: Discrete Laplace operator estimation for dynamic 3D reconstruction. In: ICCV (2019)
45. Yang, G., Ramanan, D.: Volumetric correspondence networks for optical flow. In: NeurIPS (2019)
46. Yang, G., Sun, D., Jampani, V., Vlasic, D., Cole, F., Chang, H., Ramanan, D., Freeman, W., Liu, C.: Lasr: Learning articulated shape reconstruction from a monocular video. In: CVPR (2021)
47. Yang, G., Sun, D., Jampani, V., Vlasic, D., Cole, F., Liu, C., Ramanan, D.: Viser: Video-specific surface embeddings for articulated 3D shape recon- struction. In: NeurIPS (2021)
48. Yang, G., Vo, M., Neverova, N., Ramanan, D., Vedaldi, A., Joo, H.: BANMo: Building animatable 3D neural models from many casual videos. In: CVPR (2022)

49. Yariv, L., Gu, J., Kasten, Y., Lipman, Y.: Volume rendering of neural implicit surfaces. NeurIPS (2021)
50. Yu, R., Russell, C., Campbell, N., Agapito, L.: Direct, dense, and deformable: Template-based non-rigid 3D reconstruction from rgb video. In: ICCV (2015)
51. Zhang, X., Bi, S., Sunkavalli, K., Su, H., Xu, Z.: Nerfusion: Fusing radiance fields for large-scale scene reconstruction. In: CVPR (2022)
52. Zhu, Y., Huang, D., De La Torre, F., Lucey, S.: Complex non-rigid motion 3D reconstruction by union of subspaces. In: CVPR (2014)
53. Zuffi, S., Kanazawa, A., Berger-Wolf, T., Black, M.J.: Three-D safari: Learning to estimate zebra pose, shape, and texture from images "in the wild". In: ICCV (2019)
54. Zuffi, S., Kanazawa, A., Black, M.J.: Lions and tigers and bears: Capturing non-rigid, 3D, articulated shape from images. In: CVPR (2018)